\documentclass[letterpaper, 10 pt, conference]{ieeeconf}  %

\IEEEoverridecommandlockouts    %
\let\labelindent\relax          %

\usepackage{graphics}  %
\usepackage{graphicx}
\usepackage{epsfig}    %
\usepackage{mathptmx}  %
\usepackage{times}     %
\usepackage{amsmath}   %
\usepackage{amssymb}   %
\usepackage[hidelinks]{hyperref}
\usepackage{amsthm}
\usepackage{dsfont}
\usepackage{siunitx}
\usepackage{xcolor}
\usepackage[shortlabels]{enumitem}
\usepackage{placeins}
\usepackage{dblfloatfix}
\usepackage[percent]{overpic}

\usepackage{mathtools}
\theoremstyle{definition}

\newcommand\scaleddot{\scalebox{.89}{.}}
\makeatletter
\renewcommand{\dddot}[1]{%
  {\mathop{\kern\z@#1}\limits^{\makebox[0pt][c]{\vbox to-2.2\ex@{\kern-\tw@\ex@
  \hbox{\normalfont\scaleddot\kern-0.5pt\scaleddot\kern-0.5pt\scaleddot}\vss}}}}}
\renewcommand{\ddddot}[1]{%
  {\mathop{\kern\z@#1}\limits^{\makebox[0pt][c]{\vbox to-2.2\ex@{\kern-\tw@\ex@
  \hbox{\normalfont\scaleddot\kern-0.5pt\scaleddot\kern-0.5pt\scaleddot\kern-0.5pt\scaleddot}\vss}}}}}
\makeatother

\newcommand{\x}{\mathbf{x}}

\newcommand{\ball}{\mathbf{b}}

\newcommand{\g}{\mathbf{g}}

\newcommand{\handNormal}{\mathbf{e}_{\text{h}}}

\renewcommand{\vec}[1]{\mathbf{#1}}

\newcommand{\email}[1]{\href{mailto:#1}{\texttt{#1}}}

\usepackage{soul,xcolor}
\usepackage{xparse}
\usepackage{cite}

\NewDocumentCommand{\TODO}{m g}{%
  \IfNoValueTF{#2}{%
    \sethlcolor{red!30}%
    \hl{TODO: #1}%
  }{%
    \expandafter\ifx\csname TODOcolor#1\endcsname\relax
      \sethlcolor{red!30}%
    \else
      \expandafter\sethlcolor\expandafter{\csname TODOcolor#1\endcsname}%
    \fi
    \hl{#1}\sethlcolor{red!30}\hl{-TODO: #2}%
  }%
}

\newif\ifblindreview
\blindreviewfalse

\newcommand{\FigureTopSpacing}{4.5pt}

\IfFileExists{media/rviz_legend.png}{}{
  
}

\title{\LARGE \bf Catch, Throw, Repeat: Planning for Human–Robot Partner Juggling}

\ifblindreview
\author{Anonymous\\Authors}
\else
\author{
Jonathan Rainer Lippert\textsuperscript{1},
Kai Ploeger\textsuperscript{1}, 
Abir Chowdhury\textsuperscript{2}, \\
Hermann M\"uller\textsuperscript{2},
Jan Peters\textsuperscript{1,3,4,5},
and Alap Kshirsagar\textsuperscript{1, 6}%
\thanks{
Corresponding author: \email{kai@robot-learning.de}\newline
\textsuperscript{1}Technical University of Darmstadt,
\textsuperscript{2}Justus-Liebig University Gießen,
\textsuperscript{3}German Research Center for AI (DFKI),
\textsuperscript{4}Robotics Institute Germany
\textsuperscript{5}Hessian.AI.
\textsuperscript{6}Indian Institute of Technology Delhi - Abu Dhabi.
\newline This work was supported by the Deutsche Forschungsgemeinschaft (German Research Foundation, DFG) under Germany's Excellence Strategy (EXC 3066/1 “The Adaptive Mind”, Project No. 533717223).
}}%
\fi

\begin{document}

\maketitle
\thispagestyle{empty}
\pagestyle{empty}

\begin{abstract}

Dynamic object exchange between humans and robots remains a challenging problem due to uncertainty in perception, timing, and contact-rich interaction.
Human--robot juggling represents a particularly demanding instance of this problem, requiring precise real-time coordination, predictive motion planning with feedback control, and robustness to variability in human motion.
Enabling such skills is of interest for advancing physical human--robot interaction and shared autonomy.
We present a real-time planning and control architecture for human--robot partner juggling that enables a robot to reliably catch and throw balls in synchronized multi-ball patterns with a human partner.
The system integrates predictive ball tracking,
adaptive online trajectory optimization using a multiple-shooting formulation,
and a state-machine-based coordination logic
to enable synchronized multi-ball human--robot partner juggling.
In a user study with 8 participants of varying juggling skill from beginner to expert,
we demonstrate that our system can achieve three-ball cascades shared between the robot and the human.
All participants exceeded previously reported best-case results within a 10~minute test session,
with one participant extending the previous record for shared three-ball cascade juggling fivefold to 20 consecutive robot catches,
and another participant achieving a 100\% success rate with 40 consecutive catches in a single-ball catch-and-return setting.
Video documentation can be found at
\mbox{\url{https://kai-ploeger.com/partner-juggling}} %

\end{abstract}

\section{Introduction}

Physical interaction between humans and robots often remains limited to static or quasi-static object exchange, such as scripted handovers.
In contrast, dynamic object exchange requires robots to operate under tight timing constraints while responding to uncertainty in perception, motion, and contact, often in close proximity to a human partner.
Dynamic manipulation is a key component of human dexterity and motor skill, routinely exploited in everyday tasks and skilled activities such as sports.
Enabling robots to safely and autonomously execute dynamic manipulation could increase efficiency~\cite{ha2022flingbot}, extend operational reach~\cite{zengTossingBotLearningThrow2020}, and expand interaction capabilities~\cite{masonDynamicManipulation1993}.
However, robust dynamic manipulation in unstructured, human--robot collaborative settings remains an open research problem.

\begin{figure}[t]
  \vspace{\FigureTopSpacing}
  \centering
  \includegraphics[width=\columnwidth,trim={0 45 0 75},clip]{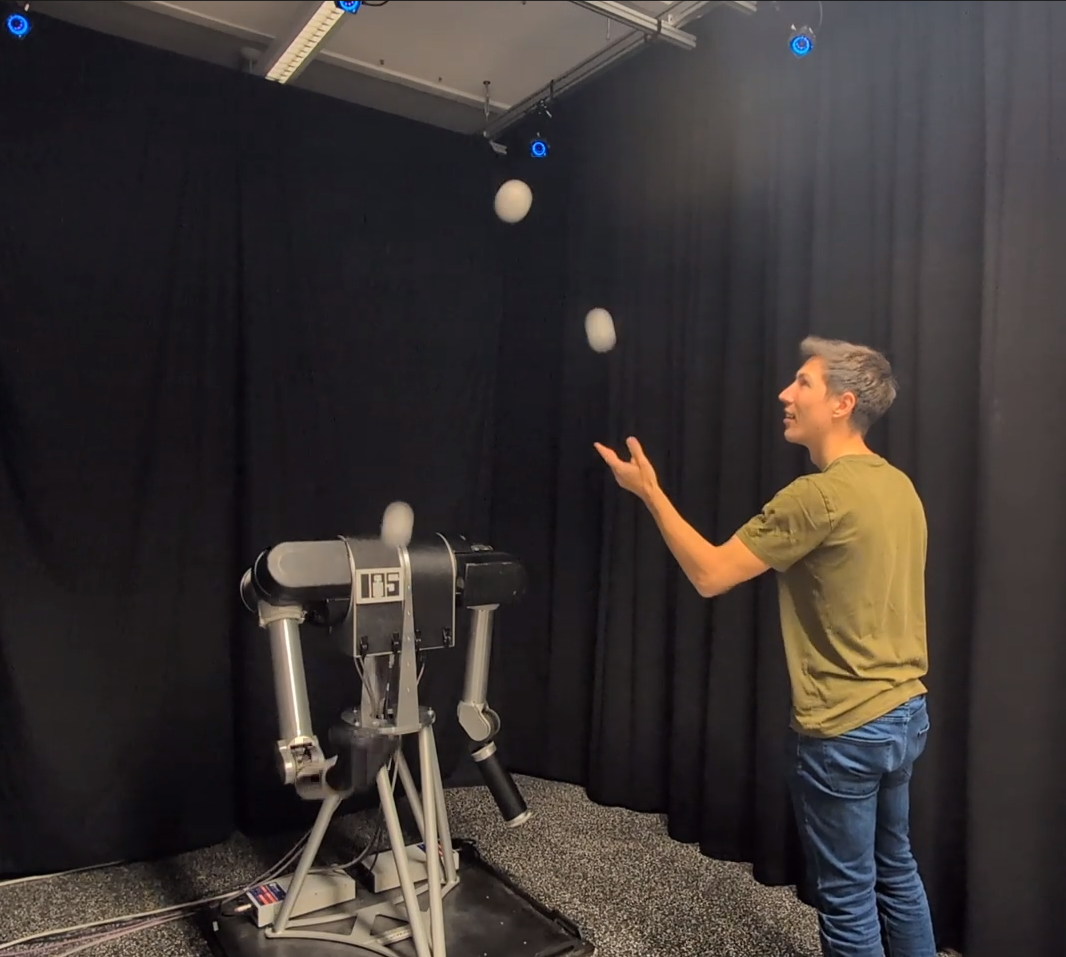}
  \caption{Human–robot partner juggling in a shared three-ball cascade. The robot and a human participant continuously exchange balls by alternately catching and throwing them, forming a synchronized 3-ball juggling pattern.}
  \label{fig:results:three_ball_cascade}
  \vspace{-15pt}
\end{figure}

Partner juggling exemplifies this class of problems.
Successful juggling requires ongoing prediction of object motion, rapid adaptation of planned trajectories, and precise feedback control, all while coordinating actions between agents.
These demands make partner juggling a challenging and scalable benchmark.
Task difficulty can be increased by varying the number of objects, and exchange patterns, while preserving the same underlying interaction structure.
In this work, we consider partner juggling tasks in which a human and a robot continuously exchange one or more balls by throwing and catching them, using one hand each, as shown in Fig.~\ref{fig:results:three_ball_cascade}.

While our approach builds directly on recent work in trajectory planning for toss juggling~\cite{ploegerControllingCascadeKinematic2022,andreuCascadeJugglingVanilla2024},
adapting to real-world human-robot partner juggling introduces several additional challenges.
In contrast to settings where all hands and throws are controlled by a single agent, the robot must adapt catch and throw timing online rather than following a predefined schedule, and continuously replan trajectories in response to changing observations.
To address these challenges, we make the following contributions.
First, we extend trajectory planning for toss juggling from simulation to a real robotic platform operating in a human--robot collaborative setting.
Second, we introduce a real-time planning and control architecture that combines predictive ball tracking, adaptive online trajectory optimization, and an explicit coordination logic for interactive catch-and-throw decisions.
Finally, we demonstrate the resulting system in a user study with participants of varying juggling skill, achieving sustained human--robot partner juggling and performance exceeding previously reported results.

\section{Related Work}

Robotic juggling and related dynamic manipulation tasks have long served as testbeds for studying accurate perception, predictive modeling, and precise control under fast object manipulation~\cite{aboaf1988task, sakaguchi1991study, ploegerHighAccelerationReinforcement2020, liu2025learning}.
Prior work spans a wide range of juggling paradigms and related subskills.
In this overview, we will focus on those involving active throwing and catching of objects
rather than batting~\cite{schaal1993open, rizzi1994further} or rolling~\cite{tanaka2021learning, woodruff2023robotic}.

\subsection{Tracking and Prediction}
Outside of exact self throws, accurate prediction of a flying object’s trajectory is essential for catching.
Prior work has employed external camera systems~\cite{namikiBallCatchingKendama2014,baumlKinematicallyOptimalCatching2010,rileyRobotCatchingEngaging2002},
motion capture systems~\cite{dongCatchBallAccurate2020,kimCatchingObjectsFlight2014},
and onboard cameras~\cite{lippiello3DMonocularRobotic2013,zhangCatchItLearning2024}.
Common prediction methods include least-squares regression~\cite{kizakiTwoBallJuggling2012,koberPlayingCatchJuggling2012},
gravity-informed fitting~\cite{schakkalDynamicObjectCatching2024},
Kalman filtering variants~\cite{namikiBallCatchingKendama2014, kimCatchingObjectsFlight2014,salehianDynamicalSystemApproach2016},
and learning-based end-to-end neural network approaches~\cite{zhangCatchItLearning2024}.

\subsection{Catching}
Robotic catching has been studied using
fixed-base manipulators~\cite{baumlKinematicallyOptimalCatching2010,koberPlayingCatchJuggling2012,kizakiTwoBallJuggling2012}
as well as mobile platforms~\cite{dongCatchBallAccurate2020,schakkalDynamicObjectCatching2024,youRunCatchDynamic2023},
with end-effector choices ranging from
nets~\cite{koberLearningThrowingCatching2012} and
simple unactuated funnels~\cite{sakaguchi1991study, burget2010visual, ploegerHighAccelerationReinforcement2020}
to fingered hands~\cite{furukawaDynamicRegraspingUsing2006, bauml2011catching, kimCatchingObjectsFlight2014},
and hybrid designs~\cite{koberPlayingCatchJuggling2012}.
Catch configurations are often determined by intersecting the predicted trajectory with a predefined plane~\cite{koberPlayingCatchJuggling2012,kizakiTwoBallJuggling2012},
or by optimizing the catch point within the reachable workspace~\cite{kimCatchingObjectsFlight2014}.
Enforcing partial or full velocity matching~\cite{uchiyamaControlRoboticManipulator2012, salehianDynamicalSystemApproach2016,andreuCascadeJugglingVanilla2024}
can avoid high-impact catches bouncing out of the end-effector and relaxes the problem of timing grasp execution,
but that is not always necessary~\cite{baumlKinematicallyOptimalCatching2010, kizakiTwoBallJuggling2012, ploegerHighAccelerationReinforcement2020}.
While individual catches have been achieved reliably, extending these approaches to dynamic multi-ball scenarios and interactive human--robot settings remains challenging.

\subsection{Throwing}
Throwing enables robots to extend their manipulable workspace. %
Most end-effector choices for throwing can be classified as either
passive funnel-shaped grippers~\cite{sakaguchi1993motion, schaal1993open, koberLearningElementaryMovements2011}
or actuated hands or grippers~\cite{senooHighspeedThrowingMotion2008, zengTossingBotLearningThrow2020, kasaei2023throwing},
with exceptions utilizing dynamic closure mechanisms~\cite{masonDynamicManipulation1993}.
Multiple planning paradigms have been proposed for throwing objects.
Most rely on kinematic models~\cite{ huBallthrowingRobotVisual2010, chenRobotThrowingTrajectory2019, andreuCascadeJugglingVanilla2024},
while~\cite{lombaiThrowingMotionGeneration2009} employ inverse dynamics to constrain peak power consumption,
and~\cite{senooHighspeedThrowingMotion2008} derive basis functions for throwing motions from a kinetic chain model of the manipulator to maximize achievable end-effector velocity.
Alternatively, a frequent approach is to define a movement primitive and tune it by hand~\cite{sakaguchi1991study, koberPlayingCatchJuggling2012, kizakiTwoBallJuggling2012}.
Removing the human feedback from the loop, learning based approaches were used to tune throwing policies through trial and error,
while sim2real transfer approaches are popular in reinforcement learning,
precise throwing policies are always at least partially trained on the real robot~\cite{aboaf1988task, koberLearningThrowingCatching2012, ghadirzadehDeepPredictivePolicy2017, ploegerHighAccelerationReinforcement2020, zengTossingBotLearningThrow2020}.
Despite this diversity of approaches, most prior work on robotic throwing assumes visually open-loop execution once the release motion is initiated.

\subsection{Full Toss Juggling Systems}

Toss juggling consists of precise throwing and reliable catching.
However, beyond solving these subskills in isolation, integrating them into continuous movements that satisfy a desired juggling pattern presents additional challenges including tight time constraints and inter-object collisions.
Prior work on toss juggling has primarily focused on single-agent scenarios,
starting with mechanical juggling automatons bouncing balls off the floor~\cite{shannonScientificAspectsJuggling1993, schaal1993open}.
Task-specific setups with two actuators per end-effector enabled demonstrations of a two-ball fountain~\cite{sakaguchi1991study, sakaguchi1993motion} and five-ball cascade~\cite{burget2010visual}, each in a planar context.
Juggling with more anthropomorphic robot arms was explored using
either funnel-shaped end-effectors~\cite{rileyRobotCatchingEngaging2002, ploegerHighAccelerationReinforcement2020}
or fingered hands~\cite{kizakiTwoBallJuggling2012, okaBallJugglingRobot2017},
achieving two-ball fountains and three-ball cascades.
To the best of our knowledge, human--robot partner juggling has only been investigated by Kober et al.~\cite{koberPlayingCatchJuggling2012}, who demonstrated short sequences of shared three-ball cascades with skilled human partners.
Taken together, existing approaches address key subskills and single-agent juggling scenarios, but leave open the problem of robust human--robot partner juggling.

\section{Method}

We approach the task of human-robot partner juggling by cascading
a ball \emph{tracking and prediction} pipeline,
and a flexible \emph{trajectory planner} capable of real-time online replanning,
using a state machine based \emph{coordination logic}.

\subsection{Ball Tracking and Prediction}

Ball positions are observed using an OptiTrack marker-based motion capture system, treating each ball as an identical point marker.
At each time step, position measurements are associated with individual ball tracks using a Hungarian assignment over Mahalanobis distances.
Each ball is tracked by an independent Kalman filter whose dynamics switch between two motion models depending on the estimated interaction phase.
During the \emph{carry phase}, when a ball is held by either the human or the robot, we use a random-walk position model,
\[
\begin{aligned}
  \vec{p}_{k+1} &= \vec{p}_k + \vec{w}^p_k, \\
  \vec{v}_{k+1} &= \vec{0},
\end{aligned}
\qquad
\vec{w}^p_k \sim \mathcal{N}\!\left(\vec{0},\,Q^p_{\text{carry}}\right),
\]
with high process noise.
During the \emph{flight phase}, when the ball is airborne, we employ a ballistic constant-acceleration model with gravity~$\vec{g}$ as the known constant acceleration,
\[
\begin{aligned}
  \vec{p}_{k+1} &= \vec{p}_k + \vec{v}_k \Delta t + \tfrac{1}{2} \vec{g}\Delta t^2 + \vec{w}^p_k, \\
  \vec{v}_{k+1} &= \vec{v}_k + \vec{g}\Delta t + \vec{w}^v_k,
\end{aligned}
\qquad
\begin{aligned}
  \vec{w}^p_k &\sim \mathcal{N}\!\left(\vec{0},\,Q^p_{\text{flight}}\right),\\
  \vec{w}^v_k &\sim \mathcal{N}\!\left(\vec{0},\,Q^v_{\text{flight}}\right).
\end{aligned}
\]
A ball is considered to be in the \emph{carry phase} when it lies within a fixed neighborhood of the robot end-effector or within a bounding region around the human partner, and in the \emph{flight phase} otherwise.
The Kalman filter state estimates are used to maintain consistent ball identities and to determine the current interaction phase via the geometric heuristic.

During the \emph{flight phase}, forward predictions required for catch planning are obtained by extrapolating a gravity-informed linear least-squares fit
\[
\begin{aligned}
  (\hat{\vec p}_0,\hat{\vec v}_0)
  &= \arg\min_{\vec p_0,\vec v_0} \sum_{i=1}^{N}
  \left\|
  \vec p_i - \Big(\vec p_0 + \vec v_0 \tau_i + \tfrac{1}{2}\vec g\,\tau_i^2\Big)
  \right\|_2^2, \\
  \hat{\vec p}(\tau)&=\hat{\vec p}_0+\hat{\vec v}_0\tau+\tfrac{1}{2}\vec g\,\tau^2.
\end{aligned}
\]
over the most recent measurements~$\vec p_i$ of the current flight phase, expressed as a function of elapsed flight time~$\tau_i = t_i - t_0$ and extrapolated forward using the predicted trajectory~$\hat{\vec p}(\tau)$.

\subsection{Adaptive Joint Trajectory Planning}
Our trajectory planning formulation builds on prior kinematic joint-space optimization approaches for robotic toss juggling~\cite{ploegerControllingCascadeKinematic2022,andreuCascadeJugglingVanilla2024}, retaining the high-level structure while replacing most constraints.
Joint trajectories are generated by solving the constrained optimization problem
\begin{equation}
\begin{aligned}
\min_{\mathbf{q}(t)} \quad & \int_{t=0}^{T} \|\ddot{\mathbf{q}}(t)\|^2 + \lambda \|\dddot{\mathbf{q}}(t)\|^2 \, dt \\
\text{s.t.} \quad & \mathbf{h}(\mathbf{x}(t), \dot{\mathbf{x}}(t), \ddot{\mathbf{x}}(t)) = 0, \\
& \mathbf{c}(\mathbf{x}(t), \dot{\mathbf{x}}(t), \ddot{\mathbf{x}}(t)) \leq 0 ,
\end{aligned}
\end{equation}
which penalizes joint accelerations~$\ddot{\mathbf{q}}(t)$ and jerks~$\dddot{\mathbf{q}}(t)$, balanced by the scalar weight~$\lambda$, to promote smooth motion, while task-specific equality constraints~$\mathbf{h}$ and inequality constraints~$\mathbf{c}$ shape the resulting end-effector motion~$\mathbf{x}(t)$ through the forward kinematic mapping $\mathbf{x}(t) = \mathbf{f}_{\mathrm{kin}}(\mathbf{q}(t))$.
All planned trajectories start at the current joint state, move to a predicted catch pose, execute a throwing motion, and follow through to recover to a safe configuration.
During the \emph{vacant phase}---while approaching the catch---the entire trajectory
is continuously replanned at up to approximately $\SI{20}{\hertz}$ using the latest available ball predictions.
The subsequent \emph{carry phase} up to release is executed open-loop to ensure consistent release conditions,
while the post-release follow-through and recovery may be canceled if a new incoming ball is detected.
In the following, we detail the individual constraints shaping this motion.

\begin{figure}[t]
  \vspace{\FigureTopSpacing} %
  \vspace{-2pt}
  \centering
    \includegraphics[trim={0 0cm 0 0cm}, clip, width=0.79\columnwidth]{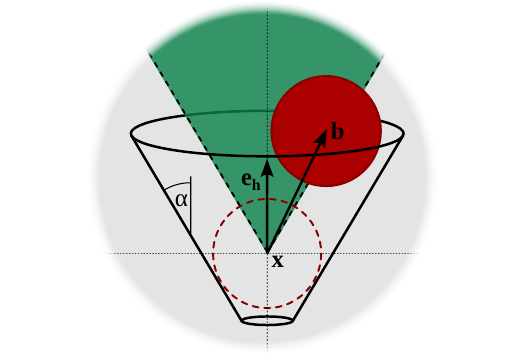}
    \vspace{-3pt}
    \caption{Post-release constraint geometry. The end-effector motion is restricted such that the predicted ball trajectory remains within the green cone-shaped region originating at the end-effector frame origin, preventing unintended contact with the ball after takeoff. During the carry phase, the ball rests at the frame origin.}
    \label{fig:method:throwing:ball-cone-draft}
    \vspace{-10pt}
\end{figure}

To \emph{catch} a ball, we intercept the predicted ball trajectory at a fixed horizontal catching plane by matching the end-effector position~$\x(t_{\mathrm{TD}})$ to the predicted touchdown location~$\tilde{\ball}(t_{\mathrm{TD}})$
with the end-effector at rest at the catch pose
\begin{equation}
\begin{aligned}
  \x(t_{\mathrm{TD}}) &= \tilde{\ball}(t_{\mathrm{TD}}), \\
  \dot{\x}(t_{\mathrm{TD}}) &= \mathbf{0}, \\
  \ddot{\x}(t_{\mathrm{TD}}) &= \mathbf{0}.
\end{aligned}
\end{equation}
We found this formulation to be more robust to slight timing uncertainties than velocity-matching constraints.
To \emph{throw} a ball, the end-effector position and velocity at release uniquely determine the subsequent ball flight, yet many release states can result in the same desired target.
To avoid overconstraining the optimization problem, we therefore constrain only the desired target position~$\ball^{*}_{\mathrm{TD+1}}$ and flight time~$T_{\mathrm{flight}}$ using a parabolic ballistic model
\begin{equation}
\begin{aligned}
  \ball^{*}_{\mathrm{TD+1}} =
  \x(t_{\mathrm{TO}}) +  \dot{\x}(t_{\mathrm{TO}}) T_{\mathrm{flight}} + \tfrac{1}{2} \g T_{\mathrm{flight}}^2,
\end{aligned}
\end{equation}
allowing the optimizer to select a dynamically feasible release state.
To ensure a clean \emph{release} at~$t_{\mathrm{TO}}$, we require the end-effector acceleration to match gravitational acceleration at the release instant and constrain the predicted ball trajectory to lie within a conical region defined by the hand geometry, as illustrated in \autoref{fig:method:throwing:ball-cone-draft}, for a short duration~$\epsilon$ after release,
\vspace{-7pt}
\begin{equation}
\begin{aligned}
  \ddot{\x}(t_{\mathrm{TO}}) &= \g, \\
  \angle\!\big(\ball(t)-\x(t), \handNormal\big) &< \alpha,
  \quad \forall\, t \in (t_{\mathrm{TO}},\, t_{\mathrm{TO}} + \epsilon].
\end{aligned}
\end{equation}
During the \emph{carry phase}, between catch and throw release, we enforce a hand-tuned minimum normal acceleration~$a_{\min}^{\perp}$ along the cone axis~$\handNormal$,
\[
(\ddot{\mathbf{x}}(t) - \mathbf{g}) \cdot \handNormal > a_{\min}^{\perp},
\quad \forall\, t \in [t_{\mathrm{TD}}, t_{\mathrm{TO}}),
\]
so that the ball remains seated inside the end-effector throughout the carry phase.
Together, these constraints ensure smooth, dynamically feasible execution of catch, throw, and release motions.

\subsection{State-machine-based Coordination Logic}

\begin{figure}[t]
  \vspace{\FigureTopSpacing} %
  \centering
  \begin{overpic}[trim={18cm 2cm 12cm 2cm},clip,width=\columnwidth]{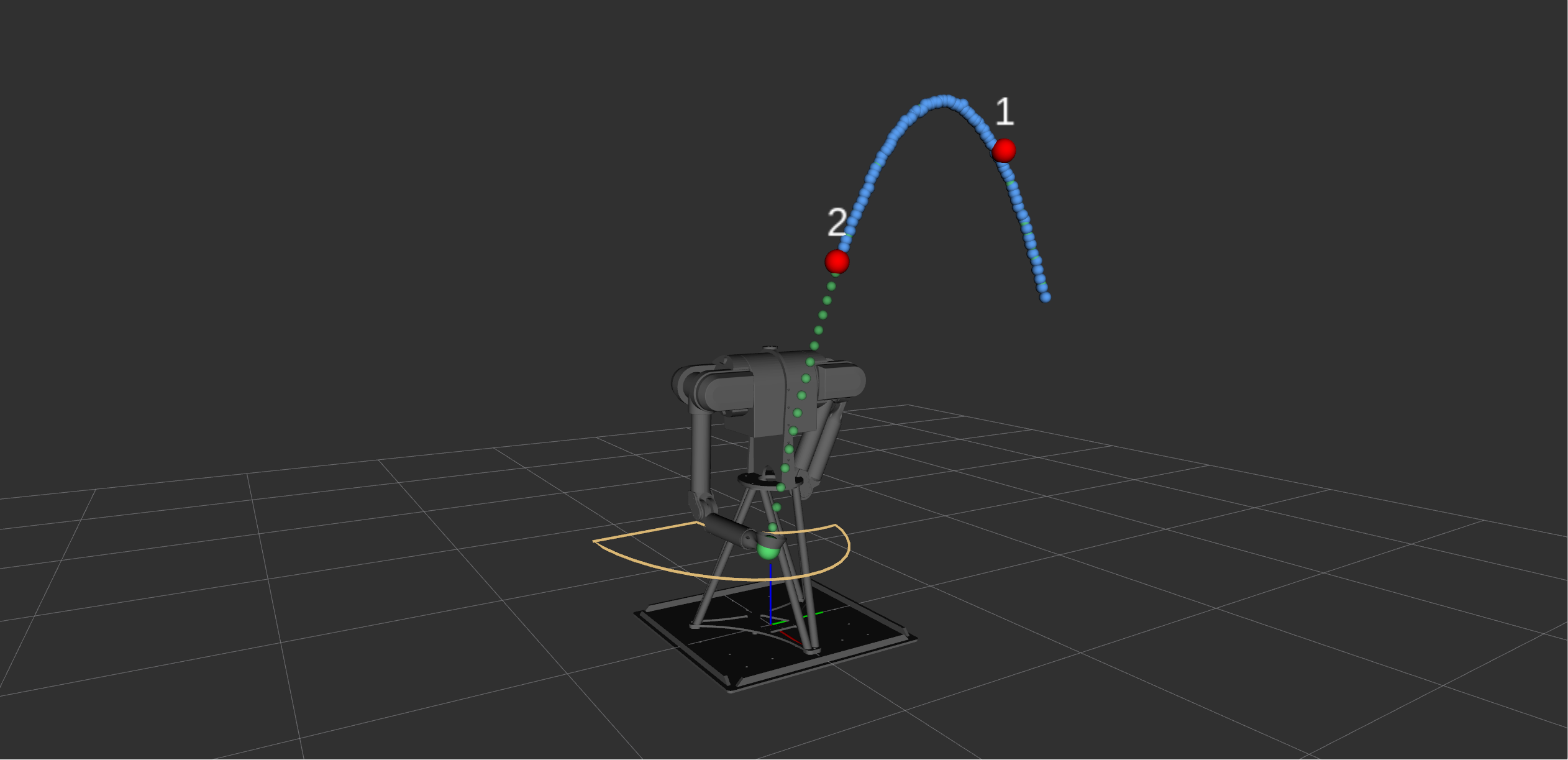}
    \put(66,3.5){\includegraphics[width=0.27\columnwidth]{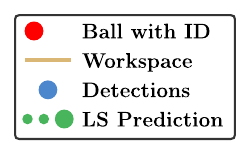}}
  \end{overpic}
  \vspace{-5pt}
  \caption{RViz screenshot of the simulated robot catching a ball within its reachable workspace in the horizontal catching plane. Touchdown locations are predicted from a least-squares fit to recent ball measurements.}
  \label{fig:method:workspace}
  \vspace{-10pt}
  \vspace{-5pt}
\end{figure}

To manage catching and throwing actions across multiple balls, we employ a discrete decision mechanism based on independent finite-state machines, one for each ball.
Simple geometric heuristics applied to the continuous ball position and velocity estimates obtained from Kalman filtering are used to cycle through the interaction phases \emph{Held by Human}, \emph{Flight to Robot}, \emph{Held by Robot}, and \emph{Flight to Human}, or \emph{Fell to Ground} if a ball is dropped.
These state labels are used to select the appropriate dynamics model for ball tracking
and to choose the ball that has been labeled as \emph{Flight to Robot} the longest to condition trajectory planning,
provided its predicted touchdown location lies within a predefined reachable workspace of the robot, illustrated in \autoref{fig:method:workspace}.
If no such ball exists, the robot remains idle and waits for a feasible incoming throw.
During the carry phase, replanning is suppressed to ensure consistent release conditions, while the post-release recovery phase may be canceled if a new incoming ball is detected,
resulting in smooth transition into the next catch movement.
This decision policy is independent of the total number of balls and therefore scales naturally to multi-ball juggling patterns.

\section{Experiments}

To evaluate the proposed approach, we conduct experiments using a $\SI{160}{\volt}$ dual-encoder Barrett WAM robotic manipulator,
configured with four degrees of freedom corresponding to the shoulder and elbow joints
and mounted in an orientation that results in anthropomorphic arm kinematics, as shown in Fig.~\ref{fig:results:three_ball_cascade}.
The cable-driven, backdrivable actuation enables accurate torque control and is well suited for dynamic catching and throwing motions.
Joint-level control runs at $\SI{500}{\hertz}$ and uses a PD controller with low uniform gains
$K_p = 150$ and $K_d = 10$ for compliance, augmented with feedforward inverse dynamics compensation to enable responsive dynamic motion.
A $\SI{170}{\milli\meter}$ diameter funnel-shaped end-effector with half opening angle~$\alpha=\ang{27}$
guides balls during catching.
We use Russian-style juggling balls (Norwik Classic) as they provide significant damping during contact.
The balls are wrapped in infrared retro-reflective tape and tracked using an OptiTrack motion capture system with eight cameras operating at $\SI{125}{\hertz}$.
All perception, planning, and control components run on a Linux workstation with a low-latency kernel (Intel i9-13900K, Ubuntu~22.04), connected to the robot via a CAN interface.
The same ROS-based interface is used for both real-world experiments and simulation, with an alternate backend based on MuJoCo~\cite{todorovMuJoCoPhysicsEngine2012}.

\vspace{-3pt}
\subsection{Human--Robot Trials}

\newcommand{\numParticipants}{8}

For the human--robot evaluation, we recruited $\numParticipants$ jugglers,
with self-reported skill levels ranging from novice three-ball cascades to proficient eight-ball fountain patterns.
Each of them was able to juggle a three-ball cascade for at least one minute on their first attempt.
Most participants had prior experience with partner juggling, primarily in the form of club passing.
Because clubs are typically caught with a pronated hand at approximately shoulder height, whereas balls are caught with a supinated hand around elbow height,
we expect limited transfer of club-passing skills to the ball-passing task studied here.
None of the participants had previously attempted partner juggling with a robot.

\begin{table}[t]
\vspace{\FigureTopSpacing}
\centering
\caption{Success rates and min-max across \numParticipants~participants.}
\vspace{-5pt}
\label{tab:results:success-streaks}
\setlength{\tabcolsep}{5pt}%
\begin{tabular*}{\columnwidth}{@{\extracolsep{\fill}}|c|c|c|c|c|@{}}
\hline
\textbf{\#Balls} & \textbf{Success (\%)} & \textbf{Max streak} & \textbf{\#Attempts} & \textbf{\#Throws} \\
\hline
1 & 87.2 (72.5--100.0) & 40 (6--40) & 320 & 320 \\
2 & 84.3 (75.0--92.0) & 35 (8--35) & 80 & 430 \\
3 & 75.1 (63.3--80.3) & 20 (5--20) & 320 & 966 \\
\hline
\end{tabular*}
\end{table}

\begin{table}[t]
\centering
\caption{Human throw metrics $\pm$ std across $\numParticipants$ participants.}
\vspace{-5pt}
\label{tab:results:throw-metrics}
\begin{tabular}{|c|c|c|c|}
\hline
\textbf{Setting} & \textbf{Period (s)} & \textbf{TD-Variability (m)} & \textbf{Height (m)} \\
\hline
1 ball (averaged) & --- & 0.184 & 1.96$\pm$0.15 \\
2 ball (averaged) & 1.95$\pm$0.2 & 0.199 & 2.02$\pm$0.19 \\
3 ball (averaged) & 1.22$\pm$0.10 & 0.240 & 2.13$\pm$0.18 \\
\hline
\hline
1 ball (pooled) & --- & 0.197 & 1.98$\pm$0.29 \\
2 ball (pooled) & 1.95$\pm$0.39  & 0.211 & 2.04$\pm$0.31 \\
3 ball (pooled) & 1.19$\pm$0.35 & 0.246 & 2.16$\pm$0.30 \\
\hline
\end{tabular}
\vspace{-8pt}
\end{table}

\begin{figure*}[t]
  \vspace{\FigureTopSpacing} %
	\centering
  \includegraphics[width=\textwidth]{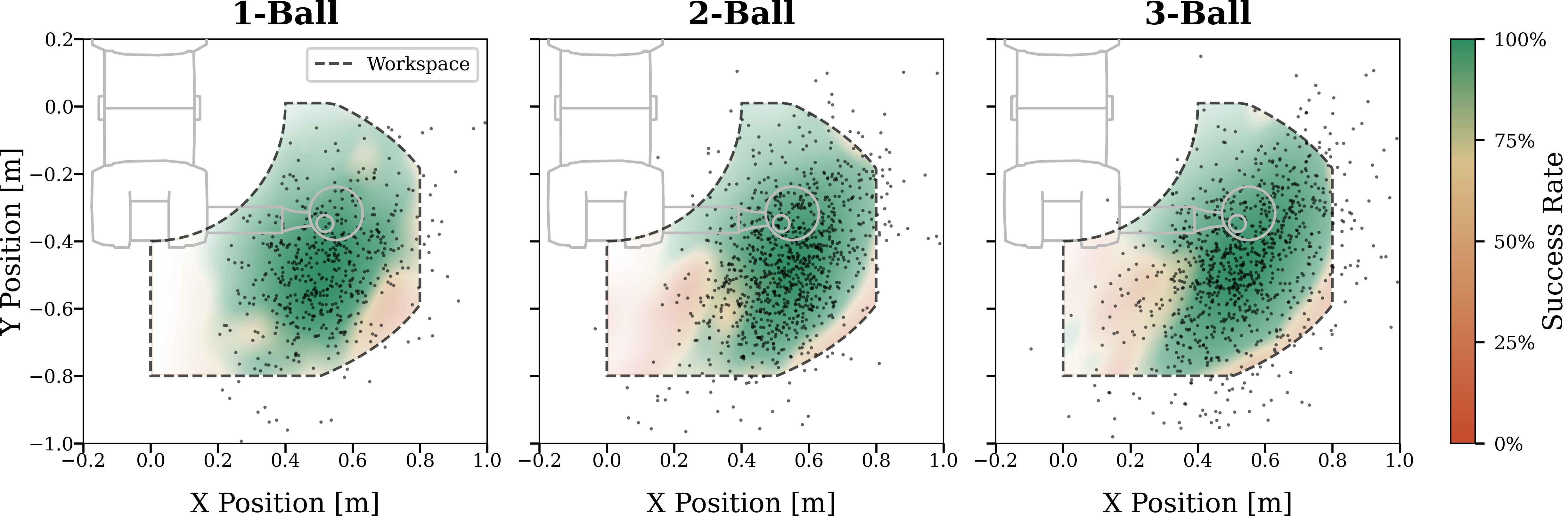}
  \vspace{-15pt}
  \caption{Top-down view of random subsamples of measured catch locations and estimated success rates.
  Successful regions cover most of the regions targeted by human throws,
  except for the rear workspace ($x < 0.3$\,m), where the arm kinematics (4-DoF) require excessive end-effector tilt. Human partners stand outside the reachable workspace, in front of and to the left of the robot (positive $x$ and $y$).}
  \vspace{-10pt}
  \label{fig:results:success-rate-over-catch-positions}
\end{figure*}

All participants followed the same \emph{experimental procedure}.
After completing a brief questionnaire on their juggling experience, participants were given five minutes to familiarize themselves with the provided juggling balls
by performing any form of solo juggling of their choice.
They were then introduced to partner juggling by attempting ten three-ball shared cascades with the experimenter as the partner.
The task subsequently proceeded with one-, two-, and three-ball trials with the robot.
For each condition, we fixed the number of attempts per participant to $50$, $20$, and $50$, and discarded the first ten to reduce warm-up and learning effects.
The two-ball setting used fewer, since our focus was the one- and three-ball patterns.
Across all \numParticipants~participants, this gives the $320$, $80$, and $320$ scored attempts in \autoref{tab:results:success-streaks}.

\vspace{\FigureTopSpacing}
As key metrics of \emph{quantitative results}, we observe the cycle success rate and the length of success streaks across conditions, summarized in \autoref{tab:results:success-streaks}.
A successful cycle is defined as a ball starting in the human hand and returning to the human hand, representing one successful catch by each partner.
Any drop, including cases where the human throw missed the robot's reachable workspace, counts as a failure for the human--robot team.
Figure~\ref{fig:results:success-rate-over-catch-positions} shows the spatial distribution of observed catch locations together with estimated success rates across one-, two-, and three-ball conditions.
Successful catches span most of the robot’s reachable workspace, indicating substantial spatial robustness despite the lack of end-effector orientation control (4-DoF robot arm).
During the ten minutes allocated to the three-ball human--robot condition, all $\numParticipants$ participants exceeded the previously reported best-case result for the 3-ball setting of four robot catches~\cite{koberPlayingCatchJuggling2012},
with one participant achieving streaks of up to 20 consecutive successful cycles.
Another participant achieved a 100\% success rate in the single-ball condition, completing 40 consecutive catch--throw cycles without a drop.
As timing constraints tighten with increasing ball count, success rates decrease and success streaks shorten, accompanied by increased variability in human throws,
as reflected in the throw metrics summarized in \autoref{tab:results:throw-metrics}.
The transition from two to three balls results in a pronounced increase in difficulty,
consistent with the well-known increase in difficulty between these patterns in human juggling.

\emph{Extended validation runs} include uninterrupted streaks of up to 28 successful cycles in the three-ball condition and 70 consecutive catch--throw cycles in the single-ball condition, with the latter ending due to a human throw missing the robot’s reachable workspace,
demonstrating substantial additional performance headroom beyond the structured participant study.

\subsection{Simulation Ablations for Mechanism Isolation}

We use a simulation to explicitly ablate failure mechanisms by systematically varying throw frequency and enabling or disabling inter-ball collisions,
both of which cannot be controlled independently in real-world human--robot partner juggling.
The complete perception, planning, and control stack used in real-world experiments is reused unchanged,
with a MuJoCo-based backend replacing robot dynamics, ball dynamics, and sensing.
Ball tracking is performed using the same tracking pipeline as on the real system,
with simulated position observations corrupted by zero-mean Gaussian noise with diagonal covariance
$\mathrm{diag}(0.002^2, 0.002^2, 0.004^2)$ to approximate OptiTrack measurement uncertainty.
Simulated control runs at $\SI{500}{\hertz}$, matching the real system, while simulated OptiTrack sensing is slightly reduced from $\SI{125}{\hertz}$ to $\SI{100}{\hertz}$ for ease of implementation.

\begin{figure}[t]
  \vspace{\FigureTopSpacing} %
  \centering
  \includegraphics[width=\columnwidth]{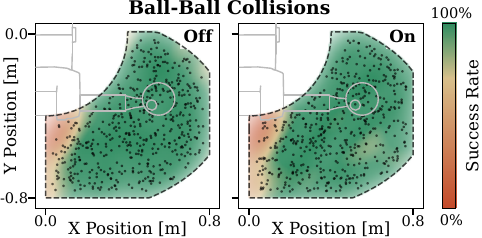}
  \vspace{-18pt}
  \caption{Top-down view of planned partner target touchdown locations and estimated robot-side success rates in simulation with and without ball--ball collisions,
  pooled over three-ball juggling frequencies.
  With collisions enabled, reduced success is observed near the center of the reachable workspace, coinciding with the robot’s typical throw release region.}
  \vspace{-15pt}
  \label{fig:results:sim-success-rate-over-workspace}
\end{figure}

In simulation, the human partner is replaced by a scripted partner model.
Partner throws are modeled by directly initializing ball states at takeoff.
For each throw, a target touchdown location is sampled uniformly from the robot’s reachable workspace, as shown in \autoref{fig:results:sim-success-rate-over-workspace}.
The ball is then launched along an exact ballistic trajectory that reaches the sampled target with a fixed flight time shared by both partner and robot throws,
i.e., partner and robot throws reach the same apex height.
Partner throws are initialized from the robot’s nominal throw target location.
Partner throws are open-loop and do not react to the robot.

Throw frequency is defined as the average number of balls thrown toward the robot per second.
Timing variability is introduced by adding zero-mean Gaussian noise with $\sigma_t^2 = 0.05^2\,\text{s}$ to the time interval between successive throws, clipped to three standard deviations via resampling.
For each frequency setting, 500 throws are simulated.
We evaluate performance across throw frequencies corresponding to partner juggling scenarios with one to four balls.

\begin{figure}[b!]
  \vspace{-5pt}
  \centering
  \includegraphics[width=\columnwidth]{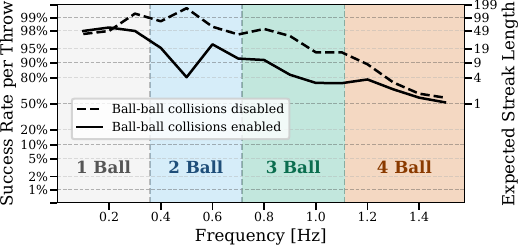}
  \vspace{-15pt}
  \caption{Robot-side success rate per throw and expected success streak length as a function of throw frequency in simulation. Disabling ball--ball collisions yields higher success rates at low and intermediate frequencies, while reduced success in the 4-ball regime limits opportunities for collisions.}
  \label{fig:results:sim-success-over-frequency}
  \vspace{-2pt}
\end{figure}

Figure~\ref{fig:results:sim-success-over-frequency} shows the robot-side success rate per throw as a function of throw frequency, together with the corresponding expected success streak length.
Each data point is obtained from a single long simulation run comprising 500 throws at a fixed mean frequency, without temporal smoothing.
Across both conditions, success rates decrease as throw frequency increases, reflecting the increasing difficulty of coordinating catch and throw actions under tighter timing constraints.
Disabling ball--ball collisions consistently improves success rates across low- and intermediate-frequency regimes.
Around the pronounced dip at $\SI{0.5}{\hertz}$, robot and partner throw simultaneously, causing opposing ballistic trajectories to cross the workspace center at the same time where their separation is minimal. This condition is well known to increase inter-ball collision probability in juggling.
At higher frequencies corresponding to four-ball juggling regimes,
success rates decrease sharply in both conditions and the performance gap between collision-enabled and collision-disabled simulations narrows due to lower overall success rates leading to fewer opportunities for collisions.

Spatial success distributions in simulation, summarized in Fig.~\ref{fig:results:sim-success-rate-over-workspace}, are stable across throw frequencies within the three-ball regime,
excluding areas shadowed by the robot body.
Each dot denotes a planned partner target touchdown location, indicating where an incoming throw is aimed.
With ball--ball collisions disabled, success rates remain uniformly high across most of the reachable workspace.
Enabling collisions does not lead to a uniform global reduction in performance.
Instead, most additional failures become strongly localized near the center of the workspace, where inbound and outbound ball trajectories overlap most frequently.
Both simulation conditions exhibit similar behavior toward the rear of the reachable workspace.
Here, reduced success persists due to the increased end-effector tilt required by the 4-DoF arm, as observed in real-world experiments,
but this effect is significantly less pronounced in simulation.

Taken together, these simulation results isolate two dominant contributors to failure: increasing timing pressure with higher throw frequencies, and localized interference due to inter-ball collisions near the robot’s throw release region.

\section{Failure Modes and Limitations}

Human--robot partner juggling is dominated by tight interaction loops between inbound and outbound motions of both partners,
which makes attribution of individual failures inherently difficult.
Small deviations in throw timing, release location, or catch execution by either partner can propagate rapidly,
leading to large downstream errors within one or two juggling cycles,
making it challenging to determine whether a failure originates from the human, the robot, or their interaction.
The simulation results presented in the previous section enable isolating such tightly coupled effects,
which cannot be disentangled reliably from real-world trials alone.

\emph{Timing pressure} constitutes a fundamental challenge of partner juggling, as observed in both real-world and simulated experiments.
As ball count increases, throw frequency rises and the temporal slack between successive interactions decreases,
reducing the available vacant phases for approaching and catching the next inbound ball.
Under these conditions, even small execution delays or tracking errors affect the entire motion,
rather than being absorbed within a single phase.
The resulting limitation of the present system is that successful operation requires a vacant phase exceeding a minimum duration~$\tau_{\min}$ between successive interactions.
When this condition is violated, the robot can no longer robustly complete the approach and catch motion before the next interaction becomes imminent.
Even when the vacant phase remains marginally above this threshold, a soft performance limit is encountered.
Shortened execution windows reduce robustness to tracking and execution delays,
leading to increased residual end-effector motion at touchdown.
Although the reference trajectories enforce zero end-effector velocity and acceleration at the planned catch time,
limited temporal slack causes the robot to effectively snatch the ball from flight,
which is inherently less reliable than intercepting a ball that has settled at a stable catch location.
Operating substantially below this threshold would require more complex strategies,
such as highly variable catch heights combined with active grasping,
or multiplex throws in which multiple balls are caught before being released jointly in a single coordinated throwing motion.
Such maneuvers are occasionally employed by human jugglers,
but lie far outside the capabilities of the current system.

\emph{Kinematic and hardware constraints} further shape the observed failure modes.
Using a four-degree-of-freedom arm configuration requires increased end-effector tilt to reach target locations near the boundaries of the reachable workspace.
This tilt reduces the effective catch funnel orientation,
leading to higher rates of bounce-offs in low-$x$ regions in our setup.
While the Barrett WAM can be configured to 7 seven degrees of freedom with an actuated wrist,
this introduces additional moving mass and dynamic coupling.
The increased mass reduces achievable release velocity,
while whipping effects in the wrist degrade throw precision.

\emph{Inter-ball collisions} constitute a second fundamental challenge of partner juggling and the dominant failure mode observed in our experiments across two- and three-ball conditions.
The current system does not explicitly reason about already inbound or outbound balls during throw planning,
nor does it coordinate intended carry or release locations with the partner.
While collision avoidance could in principle be addressed by planning around predicted ball trajectories,
even under Gaussian state estimates such as those provided by Kalman filtering,
the collision probability between two balls does not admit a closed-form solution,
making principled online optimization difficult.
As the number of balls in flight increases, the available free space for safe outbound trajectories rapidly diminishes.
Without explicit spatial coordination, the airspace becomes effectively saturated,
limiting scalability beyond two to three balls.
These observations highlight that partner juggling cannot be treated as a simple repetition of independent catch--throw cycles.
In contrast, human partners typically reduce collision risk by establishing spatial conventions for where to throw from and where to throw to,
either implicitly through interaction or explicitly through prior agreement.
Addressing inter-ball collisions therefore requires explicit coordination of throw locations,
either through predefined conventions or through online modeling of partner behavior and intent.

Importantly, several aspects of the system do not constitute primary limiting factors.
Single-ball catching and throwing are reliable across most of the reachable workspace,
and the replanning architecture operates effectively at real-time rates.
Moreover, simulated and real-world experiments exhibit qualitatively consistent failure patterns,
indicating that the dominant limitations arise from interaction dynamics and physical constraints,
rather than from deficiencies in the underlying planning or control architecture.
Finally, while the present system relies on a motion-capture setup,
which restricts experiments to controlled laboratory environments,
the proposed planning and coordination approach is not tied to this sensing modality
and could be combined with RGB or RGB-D perception in future work.

\section{Conclusions}

We presented a real-time planning and control system that enables sustained human--robot partner juggling in the real world.
In a user study with participants of varying skill, the robot reliably achieved shared one-, two-, and three-ball juggling patterns,
substantially exceeding previously reported results for human--robot partner juggling.
The approach combines predictive ball tracking, adaptive online trajectory optimization, and a state-machine-based coordination logic,
allowing robust performance under human variability and tight timing constraints.

Experimental and simulation results indicate that performance is currently constrained by timing pressure and inter-ball collisions,
which become increasingly relevant at higher ball counts.
Incorporating coordination-aware and multi-object-aware planning therefore represents a promising direction for extending partner juggling to denser and more flexible interaction patterns.

\section*{ACKNOWLEDGEMENTS}
The authors used Claude (Anthropic) for language editing and proofreading only (grammar and sentence structure). All technical content is the authors' own, and the authors take full responsibility for the final text.

\addtolength{\textheight}{-5cm}   %

\bibliography{references}

\begin{thebibliography}{10}
\providecommand{\url}[1]{#1}
\csname url@samestyle\endcsname
\providecommand{\newblock}{\relax}
\providecommand{\bibinfo}[2]{#2}
\providecommand{\BIBentrySTDinterwordspacing}{\spaceskip=0pt\relax}
\providecommand{\BIBentryALTinterwordstretchfactor}{4}
\providecommand{\BIBentryALTinterwordspacing}{\spaceskip=\fontdimen2\font plus
\BIBentryALTinterwordstretchfactor\fontdimen3\font minus
  \fontdimen4\font\relax}
\providecommand{\BIBforeignlanguage}[2]{{%
\expandafter\ifx\csname l@#1\endcsname\relax
\typeout{** WARNING: IEEEtran.bst: No hyphenation pattern has been}%
\typeout{** loaded for the language `#1'. Using the pattern for}%
\typeout{** the default language instead.}%
\else
\language=\csname l@#1\endcsname
\fi
#2}}
\providecommand{\BIBdecl}{\relax}
\BIBdecl

\bibitem{ha2022flingbot}
H.~Ha and S.~Song, ``Flingbot: The unreasonable effectiveness of dynamic
  manipulation for cloth unfolding,'' in \emph{CoRL}, 2022.

\bibitem{zengTossingBotLearningThrow2020}
A.~Zeng, S.~Song, J.~Lee, A.~Rodriguez, and T.~Funkhouser, ``{{TossingBot}}:
  {{Learning}} to throw arbitrary objects with residual physics,'' \emph{IEEE
  TRO}, 2020.

\bibitem{masonDynamicManipulation1993}
M.~Mason and K.~Lynch, ``Dynamic manipulation,'' in \emph{IEEE/RSJ IROS}, 1993.

\bibitem{ploegerControllingCascadeKinematic2022}
K.~Ploeger and J.~Peters, ``Controlling the cascade: Kinematic planning for
  {{N}}-ball toss juggling,'' in \emph{IEEE/RSJ IROS}, 2022.

\bibitem{andreuCascadeJugglingVanilla2024}
M.~G. Andreu, K.~Ploeger, and J.~Peters, ``Beyond the cascade: {{Juggling}}
  vanilla siteswap patterns,'' in \emph{{IEEE/RSJ} {IROS}}, 2024.

\bibitem{aboaf1988task}
E.~W. Aboaf, C.~G. Atkeson, and D.~J. Reinkensmeyer, ``Task-level robot
  learning,'' in \emph{IEEE ICRA}, 1988.

\bibitem{sakaguchi1991study}
T.~Sakaguchi, Y.~Masutani, and F.~Miyazaki, ``A study on juggling tasks,'' in
  \emph{IEEE/RSJ IROS}, 1991.

\bibitem{ploegerHighAccelerationReinforcement2020}
K.~Ploeger, M.~Lutter, and J.~Peters, ``High acceleration reinforcement
  learning for real-world juggling with binary rewards,'' in \emph{CoRL}, 2020.

\bibitem{liu2025learning}
Y.~Liu, B.~Da~Costa, and A.~Billard, ``Learning to throw-flip,'' in
  \emph{IEEE/RSJ IROS}, 2025.

\bibitem{schaal1993open}
S.~Schaal and C.~G. Atkeson, ``Open loop stable control strategies for robot
  juggling,'' in \emph{IEEE ICRA}, 1993.

\bibitem{rizzi1994further}
A.~A. Rizzi and D.~E. Koditschek, ``Further progress in robot juggling:
  Solvable mirror laws,'' in \emph{IEEE ICRA}, 1994.

\bibitem{tanaka2021learning}
K.~Tanaka, M.~Hamaya, D.~Joshi, F.~von Drigalski, R.~Yonetani, T.~Matsubara,
  and Y.~Ijiri, ``Learning robotic contact juggling,'' in \emph{IEEE/RSJ IROS},
  2021.

\bibitem{woodruff2023robotic}
J.~Z. Woodruff and K.~M. Lynch, ``Robotic contact juggling,'' \emph{IEEE TRO},
  2023.

\bibitem{namikiBallCatchingKendama2014}
A.~Namiki and N.~Itoi, ``Ball catching in kendama game by estimating grasp
  conditions based on a high-speed vision system and tactile sensors,'' in
  \emph{IEEE-RAS Humanoids}, 2014.

\bibitem{baumlKinematicallyOptimalCatching2010}
B.~B\"auml, T.~Wimb\"ock, and G.~Hirzinger, ``Kinematically optimal catching a
  flying ball with a hand-arm-system,'' in \emph{IEEE/RSJ IROS}, 2010.

\bibitem{rileyRobotCatchingEngaging2002}
M.~Riley and C.~G. Atkeson, ``Robot catching: {{Towards}} engaging
  human-humanoid interaction,'' \emph{Autonomous Robots}, 2002.

\bibitem{dongCatchBallAccurate2020}
K.~Dong, K.~Pereida, F.~Shkurti, and A.~P. Schoellig, ``Catch the ball:
  {{Accurate}} high-speed motions for mobile manipulators via inverse dynamics
  learning,'' in \emph{IEEE/RSJ IROS}, 2020.

\bibitem{kimCatchingObjectsFlight2014}
S.~Kim, A.~Shukla, and A.~Billard, ``Catching objects in flight,'' \emph{IEEE
  TRO}, 2014.

\bibitem{lippiello3DMonocularRobotic2013}
V.~Lippiello, F.~Ruggiero, and B.~Siciliano, ``{{3D}} monocular robotic ball
  catching,'' \emph{Robotics and Autonomous Systems}, 2013.

\bibitem{zhangCatchItLearning2024}
Y.~Zhang, T.~Liang, Z.~Chen, Y.~Ze, and H.~Xu, ``Catch it! learning to catch in
  flight with mobile dexterous hands,'' in \emph{IEEE ICRA}, 2025.

\bibitem{kizakiTwoBallJuggling2012}
T.~Kizaki and A.~Namiki, ``Two ball juggling with high-speed hand-arm and
  high-speed vision system,'' in \emph{IEEE ICRA}, 2012.

\bibitem{koberPlayingCatchJuggling2012}
J.~Kober, M.~Glisson, and M.~Mistry, ``Playing catch and juggling with a
  humanoid robot,'' in \emph{IEEE-RAS Humanoids}, 2012.

\bibitem{schakkalDynamicObjectCatching2024}
A.~Schakkal, G.~Bellegarda, and A.~Ijspeert, ``Dynamic object catching with
  quadruped robot front legs,'' in \emph{IEEE/RSJ IROS}, 2024.

\bibitem{salehianDynamicalSystemApproach2016}
S.~S.~M. Salehian, M.~Khoramshahi, and A.~Billard, ``A dynamical system
  approach for softly catching a flying object: {{Theory}} and experiment,''
  \emph{IEEE TRO}, 2016.

\bibitem{youRunCatchDynamic2023}
Y.~You, T.~Liu, X.~Liang, Z.~Xu, M.~Zhou, Z.~Li, and S.~Zhang, ``Run and catch:
  {{Dynamic}} object-catching of quadrupedal robots,'' in \emph{IEEE/RSJ IROS},
  2023.

\bibitem{koberLearningThrowingCatching2012}
J.~Kober, K.~Muelling, and J.~Peters, ``Learning throwing and catching
  skills,'' in \emph{IEEE/RSJ IROS}, 2012.

\bibitem{burget2010visual}
P.~Burget and P.~Mezera, ``A visual-feedback juggler with servo drives,'' in
  \emph{IEEE AMC}, 2010.

\bibitem{furukawaDynamicRegraspingUsing2006}
N.~Furukawa, A.~Namiki, S.~Taku, and M.~Ishikawa, ``Dynamic regrasping using a
  high-speed multifingered hand and a high-speed vision system,'' in \emph{IEEE
  ICRA}, 2006.

\bibitem{bauml2011catching}
B.~B{\"a}uml, O.~Birbach, T.~Wimb{\"o}ck, U.~Frese, A.~Dietrich, and
  G.~Hirzinger, ``Catching flying balls with a mobile humanoid: System overview
  and design considerations,'' in \emph{IEEE-RAS Humanoids}, 2011.

\bibitem{uchiyamaControlRoboticManipulator2012}
N.~Uchiyama, S.~Sano, and K.~Ryuman, ``Control of a robotic manipulator for
  catching a falling raw egg to achieve human-robot soft physical
  interaction,'' in \emph{IEEE RO-MAN}, 2012.

\bibitem{sakaguchi1993motion}
T.~Sakaguchi, M.~Fujita, H.~Watanabe, and F.~Miyazaki, ``Motion planning and
  control for a robot performer,'' in \emph{IEEE ICRA}, 1993.

\bibitem{koberLearningElementaryMovements2011}
J.~Kober and J.~Peters, ``Learning elementary movements jointly with a higher
  level task,'' in \emph{IEEE/RSJ IROS}, 2011.

\bibitem{senooHighspeedThrowingMotion2008}
T.~Senoo, A.~Namiki, and M.~Ishikawa, ``High-speed throwing motion based on
  kinetic chain approach,'' in \emph{IEEE/RSJ IROS}, 2008.

\bibitem{kasaei2023throwing}
H.~Kasaei and M.~Kasaei, ``Throwing objects into a moving basket while avoiding
  obstacles,'' in \emph{IEEE ICRA}, 2023.

\bibitem{huBallthrowingRobotVisual2010}
J.-S. Hu, M.-C. Chien, Y.-J. Chang, S.-H. Su, and C.-Y. Kai, ``A ball-throwing
  robot with visual feedback,'' in \emph{IEEE/RSJ IROS}, 2010.

\bibitem{chenRobotThrowingTrajectory2019}
H.~Chen, B.~Zhang, and T.~Fuhlbrigge, ``Robot throwing trajectory planning for
  solid waste handling,'' in \emph{IEEE CYBER}, 2019.

\bibitem{lombaiThrowingMotionGeneration2009}
F.~Lombai and G.~Szederkenyi, ``Throwing motion generation using nonlinear
  optimization on a 6-degree-of-freedom robot manipulator,'' in \emph{{{IEEE
  International Conference}} on {{Mechatronics}}}, 2009.

\bibitem{ghadirzadehDeepPredictivePolicy2017}
A.~Ghadirzadeh, A.~Maki, D.~Kragic, and M.~Bj{\"o}rkman, ``Deep predictive
  policy training using reinforcement learning,'' in \emph{IEEE/RSJ IROS},
  2017.

\bibitem{shannonScientificAspectsJuggling1993}
C.~E. Shannon, ``Scientific aspects of juggling,'' \emph{Claude E. Shannon:
  Collected Papers}, 1993.

\bibitem{okaBallJugglingRobot2017}
T.~Oka, N.~Komura, and A.~Namiki, ``Ball juggling robot system controlled by
  high-speed vision,'' in \emph{IEEE CBS}, 2017.

\bibitem{todorovMuJoCoPhysicsEngine2012}
E.~Todorov, T.~Erez, and Y.~Tassa, ``{{MuJoCo}}: {{A}} physics engine for
  model-based control,'' in \emph{IEEE/RSJ IROS}, Oct. 2012.

\end{thebibliography}
\bibliographystyle{ieeeconf}

\end{document}